%% file: root.tex
\documentclass[letterpaper, 10 pt, conference]{ieeeconf}  

\IEEEoverridecommandlockouts                              

\usepackage{svg}
\usepackage{amsmath} 
\usepackage{amssymb}  
\usepackage{xcolor}
\usepackage{algorithm}
\usepackage{algpseudocode}
\usepackage{balance}
\usepackage{wrapfig}
\usepackage{array,multirow,graphicx}
\usepackage{subcaption}
\usepackage{booktabs}
\usepackage{verbatim}
\usepackage{lipsum}
\usepackage{mathtools}
\usepackage{tikz}
\newlength{\taskimgheight}

\newlength{\sceneimgheight}
\definecolor{grey}{rgb}{0.5,0.5,0.5}
\definecolor{figred}{HTML}{DE2D2D}
\definecolor{figyellow}{HTML}{BEB703}
\definecolor{figblue}{HTML}{5CBDFD}
\algrenewcommand\algorithmicrequire{\textbf{Inputs:}}
\algrenewcommand\algorithmicensure{\textbf{Output:}}

\newif\ifnotes
\notestrue 

\ifnotes
  \newcommand{\rmm}[1]{\textcolor{blue}{RMM:~#1}}
  \newcommand{\sateesh}[1]{\textcolor{cyan}{Sateesh: #1}}

\else
  \newcommand{\rmm}[1]{}
  \newcommand{\sateesh}[1]{}

\fi

\newcommand{\methodname}[1]
{\textsc{MoMaSplat}}
\usepackage{cite} 
\makeatletter
\let\NAT@parse\undefined
\makeatother
\definecolor{mydarkblue}{rgb}{0,0.08,0.45}
\usepackage[colorlinks=true,linkcolor=mydarkblue,citecolor=mydarkblue,urlcolor=mydarkblue]{hyperref}
\usepackage{url}
\usepackage{placeins}

\title{\LARGE \bf \methodname{}: Dynamic Novel-View Demonstration Augmentation for Mobile Manipulation
}

\begin{document}

\author{Sateesh Kumar$^{*}$, Arnav Balaji$^{*}$, Arpit Bahety, Skand Peri, Georgios Pavlakos$^{\dagger}$, Roberto Mart\'in-Mart\'in$^{\dagger}$%
\thanks{$^{*}$Project lead, $^{\dagger}$Principal Investigator.}%
\thanks{The authors are with The University of Texas at Austin. {\tt\small \{sateesh, arnavbalaji21, abahety, skand, pavlakos, robertomm\}@utexas.edu}}%
}

  \maketitle
\thispagestyle{empty}
\pagestyle{empty}

\begin{abstract}

\input{sections/0_abstract}

\end{abstract}





\input{sections/1_intro}
\input{sections/2_related-works}

\input{sections/4_approach}
\input{sections/6_experiment}
\input{sections/7_conclusions}






\section*{Acknowledgment}
This work was conducted in the Robot Interactive Intelligence (RobIn) Lab at The University of Texas at Austin and was supported in part by the National Science Foundation (CAREER FAIN 2543745), the NVIDIA Academic Grant Program, Amazon, and DARPA TIAMAT program (HR0011-24-9-0428). G. Pavlakos acknowledges support by NSF-2504906 and 2544200, and the NVIDIA Academic Grant Program. S. Kumar is supported in part by the Amazon AI PhD Fellowship program. The opinions, findings, and conclusions expressed in this material are those of the authors and do not necessarily reflect the views of the sponsors.

\FloatBarrier

\bibliographystyle{IEEEtran}  
\balance  

\bibliography{references}

\clearpage
\nobalance
\appendix
\renewcommand{\thesubsubsection}{\thesubsection-\arabic{subsubsection}}
\input{sections/9_appendix}


\end{document}

%% file: sections/0_abstract.tex
Visuomotor policies trained through imitation learning are highly sensitive to changes in camera viewpoint. 
This is particularly problematic for mobile manipulation (MoMa), where robots rarely execute manipulation from the exact same base pose used during demonstration collection due to navigation uncertainty, localization drift, and environmental constraints. 
Recent approaches use novel-view synthesis (NVS) to augment demonstrations with unseen viewpoints, but they assume static scenes and therefore fail to re-render the manipulated object during the post-contact interaction phase, leaving the most critical parts of the demonstration unaugmented.
We present \methodname{}, a novel-view demonstration augmentation framework for mobile manipulation that synthesizes full manipulation trajectories from unseen robot base poses, including dynamic object interactions. 
Our key insight is that standard MoMa data collection naturally provides the multi-view observations needed to reconstruct both the static scene and the manipulated object in 3D. 
\methodname{} combines 3D Gaussian Splatting, feed-forward 3D foundation models, and kinematics-based action retargeting to separately reconstruct the scene and manipulated object, render them from novel viewpoints, and generate geometrically consistent synthetic demonstrations paired with transformed actions. 
We evaluate \methodname{} on five real-world mobile manipulation tasks involving both rigid and articulated objects across three scenes, and validate the fidelity of its synthesized demonstrations against ground truth in simulation. \methodname{} substantially improves robustness to unseen base poses, more than doubling the average success rate of the strongest baseline and maintaining non-trivial success at large base displacements where all baselines fail, while reducing the need for large-scale real-world data collection.
More information at: \url{https://robin-lab.cs.utexas.edu/momasplat/}

%% file: sections/1_intro.tex
\section{INTRODUCTION}

\label{s:intro}

Imitation learning has emerged as a powerful paradigm for training visuomotor policies that perform complex manipulation tasks directly from image observations~\cite{chi2023diffusionpolicy, zhao2023learning}. 
However, these policies are notoriously sensitive to changes in camera viewpoint~\cite{yang2025mobi, heo2026anycamvla, fei2025libero}. 
Even small deviations in camera pose can significantly degrade performance, since the policy learns a narrow visual distribution tied to the demonstrations used during training. 

This sensitivity is particularly problematic for mobile manipulation (MoMa). 
Unlike tabletop manipulation setups, where demonstrations are often collected from a fixed robot camera pose, mobile manipulators rarely execute tasks from the exact same location at deployment time. 
Navigation uncertainty, localization drift, obstacle avoidance, and imperfect planning routinely place the robot at different base poses relative to the workspace. 
As a result, the robot observes the manipulation scene from viewpoints that may differ substantially from those seen during demonstration collection. 
Robust deployment therefore requires visuomotor policies that generalize across a distribution of robot base poses rather than a single viewpoint.

A straightforward solution would be to collect demonstrations from many different robot locations. 
However, large-scale real-world teleoperation is expensive and difficult to scale~\cite{fu2024mobile, zhao2023learning, mandlekar2023mimicgen}. 
Recent work has therefore explored synthesizing additional demonstrations through novel-view synthesis (NVS), generating virtual observations from unseen viewpoints using real-world captures~\cite{pan2025one, yang2025novel, zhou2023nerf}. 
While promising, existing approaches assume a static scene and therefore fail during the most important part of manipulation: object interaction. 
Since the manipulated object moves throughout the demonstration, prior NVS methods cannot consistently re-render the object from novel viewpoints during the post-contact interaction phase, restricting augmentation primarily to pre-contact motions or introducing geometrically inconsistent generations.

\begin{figure}
    \centering
    \includegraphics[width=\linewidth]{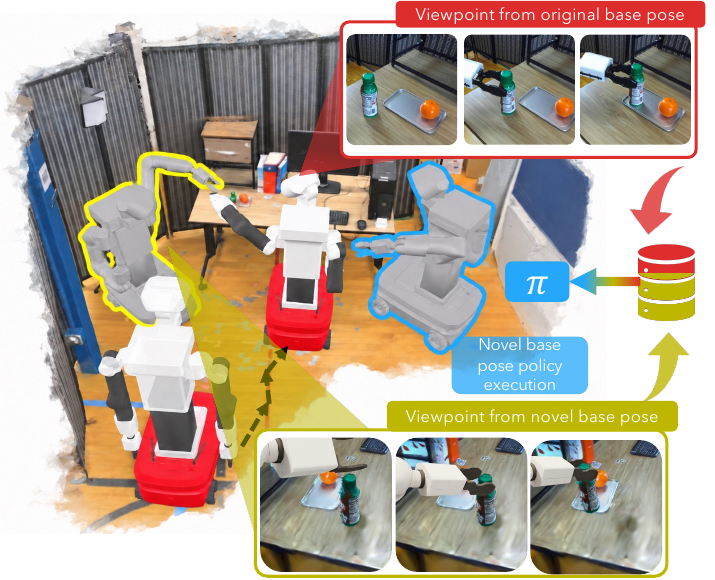}
     \caption{\textbf{\methodname{}.}
Starting from mobile-manipulation data collected at the original base pose (white and \textcolor{figred}{red} robot), \methodname{} synthesizes demonstrations from sampled novel base poses (\textcolor{figyellow}{yellow}-outlined gray robot). The top-right sequence shows the source manipulation demonstration, while the bottom-right shows the corresponding synthesized trajectory, including the dynamic object as it is moved by the robot. Augmenting the source manipulation dataset with these demonstrations enables execution from unseen deployment-time base poses (\textcolor{figblue}{blue}-outlined gray robot).}
     
    \label{fig:teaser}
\end{figure}

In this work, we present \methodname{}, a demonstration augmentation framework for mobile manipulation that synthesizes complete manipulation trajectories from unseen robot base poses, including dynamic object interactions (Fig. \ref{fig:teaser}). 
Our key insight is that standard mobile-manipulation data collection naturally provides the multi-view observations needed to reconstruct both the static scene and the manipulated object in 3D. 
During navigation, the robot observes the scene from diverse viewpoints that enable scene reconstruction, while the manipulation demonstrations themselves provide object-centric observations that enable reconstructing the manipulated object and its motion.

Building on this observation, \methodname{} separately reconstructs the static scene and manipulated objects using 3D Gaussian Splatting~\cite{kerbl20233d} and feed-forward 3D foundation models~\cite{wang2025vggt}. 
Given a novel robot base pose, \methodname{} re-renders the scene, manipulated object, and robot from the new viewpoint and retargets the original actions through kinematics-based transformations to generate geometrically consistent synthetic demonstrations.

\noindent We evaluate \methodname{} on five real-world mobile manipulation tasks involving both rigid and articulated objects across three scenes, and validate the fidelity of its synthesized demonstrations against ground truth in simulation.
Our experiments show that \methodname{} substantially improves robustness to unseen deployment-time base poses, outperforming prior augmentation approaches across $12$ unseen base poses ($3$ rollouts per pose) per task as the robot moves farther away from the source demonstration locations. 

\noindent Our contributions are threefold:
\begin{itemize}
\item A novel-view demonstration augmentation framework for mobile manipulation that synthesizes complete trajectories, including dynamic object interactions.
\item A geometry-consistent generation pipeline that separately reconstructs scenes and manipulated objects and retargets actions to unseen base poses.
\item Substantial improvements in robustness to deployment-time viewpoint variation across multiple real-world mobile manipulation tasks.
\end{itemize}

%% file: sections/2_related-works.tex
\section{Related Work}
\label{sec:related}

Our method proposes a solution to train imitation learning visuomotor policies for mobile manipulators via data augmentation.
Here, we review the most relevant related work in those areas.

\textbf{Demonstration Synthesis in Simulation.} A prominent line of work synthesizes new demonstrations by replaying source actions in randomized \textit{simulated} scenes. The X-Gen family~\cite{mandlekar2023mimicgen, jiang2025dexmimicgen, garrett2024skillmimicgen, li2025momagen, lin2025constraint} procedurally generates data by transforming and sequencing object-centric trajectory segments, with extensions such as MoMaGen~\cite{li2025momagen} addressing mobile-manipulation constraints like base reachability and visibility. However, because the generated plans are purely kinematic, they typically rely on open-loop on-robot rollouts to verify feasibility and capture visual observations, at a cost that rivals manual data collection~\cite{xue2502demogen, pan2025one}. A parallel line of work uses simulators to generate data from multiple viewpoints~\cite{sadeghi2018sim2real, hirose2022exaug, seo2023multi}, but requires modeling the full environment and task dynamics in simulation, suffering from the visual and physical sim-to-real gap. Our work avoids on-robot rollouts and full-scene simulation by reconstructing the real scene and manipulated object via novel-view synthesis, using simulation only to render the known robot embodiment.

\textbf{Demonstration Augmentation from Real-World Captures.} Beyond 2D image augmentation~\cite{laskin2020reinforcement, yarats2021image, hansen2020self}, recent methods synthesize observations from real-world captures. Neural simulators~\cite{jang2025dreamgen, bruce2024genie, agarwal2025cosmos} use generative video models but can suffer from hallucinations and geometric inconsistencies. Alternatively, novel-view synthesis (NVS) enables viewpoint augmentation~\cite{yang2025novel, zhou2023nerf, xue2502demogen, tian2024view, tagliabue2023tube, chen2024rovi, wang2026efficient, zhang2024diffusion}. VISTA~\cite{tian2024view} uses ZeroNVS~\cite{sargent2024zeronvs}, while Yang et al.~\cite{yang2025novel} employ 3D Gaussian Splatting~\cite{kerbl20233d}; however, both focus on third-person tabletop settings requiring multi-camera pre-scans. Closer to our setting, One-Thousand Trajectories~\cite{pan2025one} supports moving eye-in-hand cameras, while EgoDemoGen~\cite{xu2025egodemogen} generates paired egocentric observations and actions using a generative model. However, these methods do not explicitly reconstruct dynamic objects and restrict augmentation to pre-contact motion. In contrast, \methodname{} jointly models the static scene and manipulated objects, enabling kinematically consistent augmentation across both pre-contact and post-contact interaction phases for mobile robots with egocentric cameras.


\textbf{Mobile Manipulation.} Mobile manipulation remains a key challenge in robot learning~\cite{bahety2025safemimic, wu2023tidybot, liu2024okrobot, jiang2025wbvima, yenamandra2024homerobot}, particularly in selecting robot base poses. Traditional methods use geometric heuristics such as Inverse Reachability Maps~\cite{stulp2012learning, jauhri2022robot, rosen2023synthesizing}, while learning-based navigation~\cite{hirose2024lelan, yokoyama2024vlfm, yokoyama2023asc, shafiullah2023bringing} often overlooks downstream policy sensitivity to viewpoint changes. Mobi-$\pi$~\cite{yang2025mobi} selects base poses matching the training distribution but requires pre-scanning and test-time optimization. In contrast, we train policies robust to diverse base poses, reducing the need for precise base placement at deployment.

%% file: sections/4_approach.tex
\section{\methodname{}}

We aim to augment a mobile-manipulation dataset with synthetic demonstrations from unseen robot base poses. From source demonstrations at a few robot locations, \methodname{} synthesizes observations and actions for nearby deployment-time poses never physically demonstrated.

To achieve this, \methodname{} decomposes the environment into three components: i) a static scene model, ii) a dynamic object model for the manipulated object, and iii) the robot embodiment itself. 
The scene and object models are reconstructed independently and aligned into a shared metric frame. 
A key observation is that the navigation trajectories naturally provide the bridge between both reconstructions: during navigation, the manipulated objects remain static, allowing the object-centric and scene-centric trajectories to be consistently registered.

\begin{figure*}[t]
    \centering
    \vspace*{0.5\baselineskip}
      \includegraphics[width=1.0\textwidth]{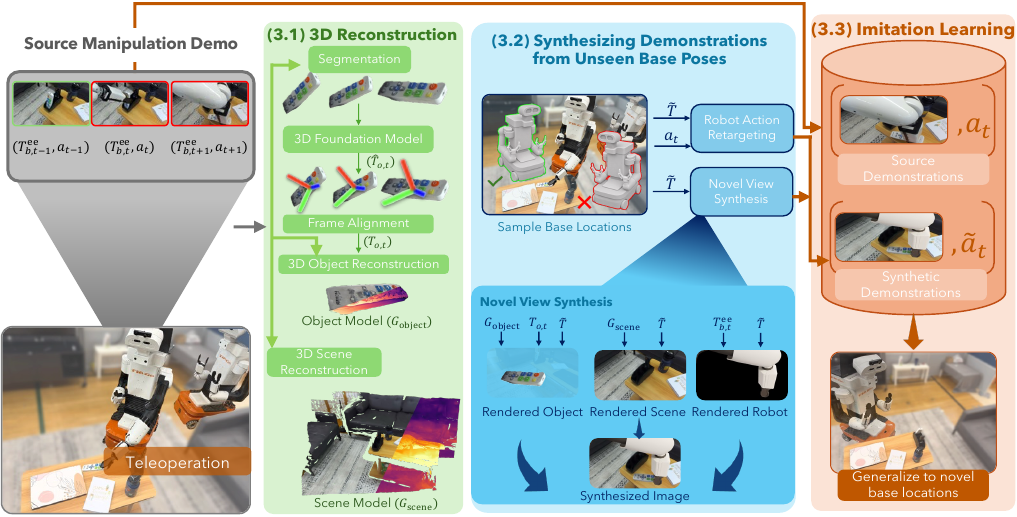}
    \caption{\textbf{Overview of \methodname{}.} \emph{3D Reconstruction:} We build 3D models of the environment from the navigation and manipulation data. We estimate camera poses via structure-from-motion and fit a 3D Gaussian Splatting model of the static scene, $\mathcal{G}_{\text{scene}}$. For the dynamic objects, we segment each manipulated object and estimate its per-frame pose trajectory with a feed-forward 3D foundation model; since the scene and object poses lie in separate reference frames, we \emph{align} them with a Sim(3) transformation and fit a separate object model $\mathcal{G}_{\text{obj}}$. \emph{Synthesizing Demonstrations from Unseen Base Poses:} Given a robot base pose, $\tilde{T}$, sampled around the original demonstration poses (left), we render the scene and object from the new viewpoint, transform the source end-effector trajectory into the new base frame to render the robot from a simulator, and composite the three streams into a photorealistic novel observation. Paired with the transformed actions $\tilde{a}_{1:L_j}$, this forms a synthetic demonstration; repeating across sampled poses yields the generated dataset. \emph{Imitation Learning:} the policy is trained on the union of source and generated demonstrations.}
    \label{fig:pipeline}
\end{figure*}

For augmentation, given a sampled robot base pose, \methodname{} renders the scene, manipulated object, and robot from the corresponding viewpoint and retargets the demonstrated actions through kinematic transformations. The resulting observations and transformed actions form synthetic demonstrations that augment the dataset for policy training. Fig.~\ref{fig:pipeline} provides an overview of the full process.

\noindent \textbf{Problem Formulation.} We consider an imitation learning setting on a mobile manipulator operating in a scene. 
The robot alternates between two modes of operation: navigation, which changes the robot base pose, and manipulation, which executes arm actions from a fixed base location. 
We represent the robot base pose as a transform ${}^{b}_{w}T \in SE(2)$ with respect to a stationary world frame $w$, parameterized by a planar position $(x_b, y_b)$ and heading $\theta_b$. 
The robot observes the scene through an egocentric RGB camera mounted on the robot body (e.g., head or torso). 
The camera pose relative to the robot base is determined by the robot configuration through forward kinematics, such that both changes in base pose and articulation of the robot body induce changes in viewpoint.

During data collection, the robot navigates to a predefined set of $N$ target base poses:
\[
\mathcal{T} = \{{}^{b}_{w}T_1, \dots, {}^{b}_{w}T_N\},
\]
from which manipulation demonstrations are collected. 
The robot executes $N_{\text{nav}}$ navigation trajectories from different starting locations toward the target base poses, either teleoperated or autonomously, and performs $M$ manipulation demonstrations from each fixed base pose without further base motion.

This process produces two complementary datasets. First, we collect a navigation dataset $\mathcal{D}_{\text{nav}} = \{d^1_{\text{nav}}, \dots, d^{N_{\text{nav}}}_{\text{nav}}\}$, where $N_{\text{nav}}$ denotes the number of navigation trajectories and each trajectory $d^k_{\text{nav}} = \big((I_1, q_1), \dots, (I_{L_k}, q_{L_k})\big)$ contains RGB observations and robot states recorded while navigating toward one of the target base poses. These trajectories provide diverse multi-view observations of the environment and manipulated objects before interaction. Second, we collect a source manipulation dataset $\mathcal{D}_{\text{src}} = \{d^1_{\text{src}}, \dots, d^{N_{\text{src}}}_{\text{src}}\}$ of $N_{\text{src}} = N \times M$ demonstrations, where each trajectory $d^j_{\text{src}} = \big((I_1, q_1, a_1), \dots, (I_{L_j}, q_{L_j}, a_{L_j})\big)$ contains manipulation observations and actions recorded during task execution from one of the $N$ base locations.
We represent actions as 7-D commands $a_t = (p_t, o_t, g_t)$, consisting of desired end-effector translation and rotation deltas $p_t \in \mathbb{R}^3$ and $o_t \in SO(3)$ applied in the robot base frame, together with the gripper command $g_t \in \mathbb{R}$.

Our goal is to augment the source manipulation dataset $\mathcal{D}_{\text{src}}$ with a synthetic dataset $\mathcal{D}_{\text{gen}}$ of demonstrations generated at unseen base poses 
to obtain a base-pose robust policy trained on $\mathcal{D} = \mathcal{D}_{\text{src}} \cup \mathcal{D}_{\text{gen}}$. To achieve this, \methodname{} first reconstructs the scene and manipulated objects in 3D and aligns them into a shared metric frame (Sec.~\ref{sec:3d_reconstruction}). It then synthesizes demonstrations from sampled base poses by rendering the scene, object, and robot from the corresponding viewpoint and retargeting the actions and proprioception to remain geometrically consistent (Sec.~\ref{sec:demo_synthesis}). The synthesized demonstrations are combined with the source dataset for policy training (Sec.~\ref{sec:imitation_learning}).

\subsection{3D Reconstruction}
\label{sec:3d_reconstruction}
To synthesize demonstrations from unseen robot base poses, \methodname{} requires a 3D representation that can be rendered from arbitrary viewpoints. A central challenge is that manipulation involves dynamic objects whose pose changes throughout the demonstration. Unlike prior novel-view synthesis approaches that model the environment as a single static scene, \methodname{} reconstructs the static scene and manipulated objects separately, leveraging observations naturally produced during mobile-manipulation data collection. Navigation
trajectories cover the environment while the objects are static, and manipulation trajectories
capture the objects in motion.

\noindent
\textbf{3D Scene Model Reconstruction.}
\label{sec:scene_model_fitting}
As the robot moves through the workspace, its 
camera naturally captures diverse viewpoints, providing the multi-view coverage needed for 3D reconstruction without external cameras or scanning. 
We therefore use $\mathcal{D}_{\text{nav}}$ to build a photorealistic 3D scene model: we recover per-image camera poses with structure-from-motion via COLMAP~\cite{schonberger2016structure}, which also produces a sparse point cloud that initializes a 3D Gaussian Splatting (3DGS)~\cite{kerbl20233d} reconstruction.

We optimize the Gaussians by minimizing a photometric loss between rendered and observed images. Since the scene is not perfectly static across navigation trajectories, transient content may appear between demonstrations; we adopt the uncertainty-aware formulation of WildGaussians~\cite{kulhanek2024wildgaussians}, which down-weights inconsistent regions during optimization to yield a robust model $\mathcal{G}_{\text{scene}}$. We additionally mask out pixels belonging to the manipulated object and the robot during scene reconstruction, preventing the object from being baked into $\mathcal{G}_{\text{scene}}$ and avoiding a static ``ghost'' when the independently reconstructed dynamic object is later composited into the scene.

Structure-from-motion reconstructs the scene only up to an unknown global scale, while our framework transforms metric actions and therefore requires a metric reconstruction. We recover the scale using the robot's onboard depth sensor: for the frames in $\mathcal{D}_{\text{nav}}$, we compare the measured depth maps to depth maps rendered from $\mathcal{G}_{\text{scene}}$ and take the median per-pixel ratio of measured to rendered depth as the scale factor.
\noindent
\textbf{3D Object Model Reconstruction.}
\label{sec:object_model_fitting}
Since the object of interest moves independently from the scene, we build a separate 3DGS model for it. 
Conceptually, the steps mirror those for the scene: we estimate the object's pose in each frame and use these poses to optimize a 3D Gaussian reconstruction. 
However, the challenges here are different.
Correspondence-based methods~\cite{schonberger2016structure} are poorly suited to small, weakly-textured objects,
%
while instance-level pose trackers~\cite{wen2024foundationpose} 
require an existing 3D model of the object, which is not available to us. 
Instead, we find that our setting is amenable to recent feed-forward 3D foundation models, like VGGT~\cite{shen2025fastvggt, wang2025vggt}, which estimate camera poses directly from an unposed RGB sequence. 



Specifically, for each trajectory $d^k_{\text{nav}} \in \mathcal{D}_{\text{nav}}$ and source demonstration $d^j_{\text{src}} \in \mathcal{D}_{\text{src}}$, we extract the RGB frames $\{I_t^k\}_{t=1}^{L_k}$ and $\{I_t^j\}_{t=1}^{L_j}$, respectively, and obtain per-frame segmentation masks $\mathcal{M}_t^k$ and $\mathcal{M}_t^j$ for the object using an open-world segmentation pipeline~\cite{ren2024grounded, ravi2024sam2}, which masks out the background and the robot.

Then, we apply FastVGGT~\cite{shen2025fastvggt} to the object-masked frames $\{I_t^k \odot \mathcal{M}_t^k\}$ and $\{I_t^j \odot \mathcal{M}_t^j\}$, respectively.
%
With everything else masked out, 
the model predicts a camera pose at each frame that is expressed relative to the object itself. 
These object-relative camera poses $\hat{T}_{o,t} \in SE(3)$ act as the initial pose estimates in the model's internal coordinate frame $v$.

\textbf{Frame Alignment.} The estimates $\hat{T}_{o,t}$ lie in $v$, with arbitrary scale, orientation, and origin relative to the metric world frame $w$ established by the Scene Reconstruction.
To align the two frames, we use a set of frames from the navigation demonstrations $\mathcal{D}_{\text{nav}}$ as a bridge. During navigation the object remains static, so the only motion comes from the robot itself, i.e., the pose trajectory recovered by the object model is purely due to camera motion. This is not true for the manipulation frames: the object moves, so the recovered trajectory conflates camera and object motion. On the selected navigation frames, the same cameras are posed twice, by the feed-forward model in $v$ and by structure-from-motion in $w$. The two trajectories must describe the same motion up to an unknown Sim(3) transformation.
We recover this transformation $\mathbf{S}: v \to w$ in closed form via the Umeyama algorithm~\cite{umeyama1991least}. Applying $\mathbf{S}$ to the estimated poses yields the aligned pose trajectory in the world frame, $T_{o,t} = \mathbf{S} \cdot \hat{T}_{o,t}$.

\textbf{Object 3DGS Optimization.} Finally, we jointly optimize these aligned poses $T_{o,t}$ alongside the Gaussian parameters of $\mathcal{G}_{\text{obj}}$, backpropagating the photometric loss through both.
This yields a pose-disentangled representation $\mathcal{G}_{\text{obj}}$ that can be rendered at arbitrary 6-DoF poses for compositing into novel-pose observations.
For articulated objects, we model the moving rigid link separately from the static remainder of
the object. For example, a laptop lid or the basket of an air fryer is represented by $\mathcal{G}_{\text{obj}}$, while the keyboard base or the air fryer's surrounding body remains
part of the static scene. The component represented by $\mathcal{G}_{\text{obj}}$ therefore
undergoes a rigid 6-DoF motion over the demonstrated trajectory, and we do not estimate
articulation parameters or extrapolate beyond the demonstrated motion.

\subsection{Synthesizing Demonstrations from Unseen Base Poses}
\label{sec:demo_synthesis}
Given the scene and object models, our framework synthesizes new demonstrations at unseen base poses in three stages. For each original robot base pose, $^\mathit{b}_\mathit{w}{T}_i$, we first sample a set of target base poses that cover possible deployment-time locations while remaining kinematically feasible (e.g., reachable for the arm). 
We then synthesize each source demonstration from each sampled pose by composing three separately rendered streams: the static scene (from the 3D scene model), the manipulated objects (from the 3D object model), and the robot body as seen from the new base location.
Finally, we retarget the source action to be defined with respect to the virtual sampled base location to obtain a new pair of virtual image and action label.

\textbf{Sampling Base Poses.}
\label{sec:camera_sampling}
To generate synthetic demonstrations, we sample a set of target base poses $\mathcal{T}^{\text{sampled}}_i=\{\tilde{T}_1, \ldots, \tilde{T}_{K}\}$ (here, we omit the superindex $b$ and subindex $w$ from $^\mathit{b}_\mathit{w}\tilde{T}$ for simplicity) that the policy should learn to handle around each of the original base locations, ${T}_i$. 
We sample the planar position $(\tilde{x}_j, \tilde{y}_j)$ uniformly within a region surrounding the original base pose, chosen to cover the range of deployment-time poses the robot may reach. 
For each sampled position, we set the base heading $\tilde{\theta}_j$ so that the object of interest is centered in the egocentric camera's field of view, then perturb it with uniform noise to expose the policy to a range of object-in-frame configurations. 
We then apply the transformation $\Delta ^i_jT = (\tilde{T}_j)^{-1} \cdot {T}_i$ that transforms the original base frame, ${T}_i$, to the sampled base frame, $\tilde{T}_j$ (where we lift $\Delta ^i_jT$ from $SE(2)$ into $SE(3)$ with zero vertical translation and zero roll/pitch when transforming 3D poses and vectors).
We retain a sampled base pose, $\tilde{T}_j$, only if the source arm trajectory, where the end-effector remains at the same pose with respect to the manipulated object, admits a valid inverse-kinematics solution at several probe timesteps spanning the demonstration; infeasible poses are rejected and resampled. 
To obtain the end-effector pose in the sampled base frame, we apply $\Delta_j^i T$ to the source end-effector pose
\ensuremath{{}^{ee}_{b}T_i}, obtained from \ensuremath{q} through
forward kinematics. The resulting pose,
\ensuremath{{}^{ee}_{b}\tilde{T}_j = \Delta_j^i T \cdot {}^{ee}_{b}T_i},
preserves the end-effector pose in the world frame.

\textbf{Novel View Synthesis.}
\label{sec:view_synthesis}
We virtually place the egocentric camera at its mounted offset relative to $\tilde{T}_j$ using forward kinematics and the robot's recorded configuration, $q$, and matching the physical camera's intrinsics.
We then render three elements from the new view: i) the scene, ii) the manipulated object, and iii) the robot. 
From the scene model, $\mathcal{G}_{\text{scene}}$, 
we obtain a virtual RGB image $\tilde{I}_{\text{scene}}$, the depth map $\tilde{D}_{\text{scene}}$ and the accumulated opacity $\tilde{\alpha}_{\text{scene}}$. 
For the manipulated object, we transform $\mathcal{G}_{\text{obj}}$ to its source-demonstration pose $T_{o,t}$
and rasterize it from the same camera, yielding per-timestep renders $\tilde{I}_{\text{obj}}$, depth maps $\tilde{D}_{\text{obj}}$ and opacity image $\tilde{\alpha}_{\text{obj}}$.
For the robot, we use inverse kinematics on the end-effector pose for the sampled base location, $^{ee}_b\tilde{T}_j$, to recover a joint configuration, $\tilde{q}^j_t$, that would place the end-effector at the same location with respect to the object. Then, we apply the joint configuration to the robot and render it using OmniGibson~\cite{li2023behavior} to obtain a robot-only RGB image $\tilde{I}_{\text{robot}}$, depth map $\tilde{D}_{\text{robot}}$ and opacity mask $\tilde{\alpha}_{\text{robot}}$.

To get a single RGB observation from the three virtual images, we need to combine them together, while respecting the depths and the opacities at each pixel.
To achieve this, we use per-pixel alpha compositing.
Specifically, at each pixel, we sort the three layers by 
depth, letting $(c_i, \alpha_i)$ denote the color and opacity of  the $i$-th layer, with $i=1$ being the nearest.
%
Then, we combine them in a front-to-back manner:
    $c = \sum_{i=1}^3 c_i \alpha_i \prod_{j=1}^{i-1} (1 - \alpha_j)$,
so that nearer surfaces occlude farther ones.
%
This generates virtual RGB images as would be seen from a physical camera at sampled location $\tilde{T}_j$.

\textbf{Robot Action and Proprioception Retargeting.}
A change in base pose moves the coordinate frame in which both the robot's end-effector proprioception and actions are expressed, so both must be retargeted to the sampled base frame $^\mathit{b}_\mathit{w}\tilde{T}_j$.
Because the policy conditions on end-effector proprioception in the robot base frame (consisting of the end-effector pose ${}^{ee}_{b}T_i$ and gripper state), we transform the end-effector pose into the sampled base frame via ${}^{ee}_{b}\tilde{T}_j = \Delta^i_j T \cdot {}^{ee}_{b}T_i$ (as above) while keeping the gripper state unchanged.
Robot actions $a_t = (p_t, o_t, g_t)$ consist of desired end-effector translation deltas $p_t \in \mathbb{R}^3$, rotation deltas $o_t \in SO(3)$, and gripper commands $g_t \in \mathbb{R}$, where both translation and rotation deltas are defined and applied in the robot base frame (i.e., ${}^{ee}_{b}R_{t+1} = R(o_t)\,{}^{ee}_{b}R_t$).
To retarget the desired end-effector translation delta, we apply the rotation component of $\Delta ^i_jT$ so that $\tilde{p}_t=R_{\Delta ^i_jT} \cdot p_t$ expresses the same world-frame displacement in the sampled base frame.
Because rotation deltas are applied in the base frame, conjugating $R(o_t)$ by $R_{\Delta ^i_jT}$ rotates its axis of rotation by $R_{\Delta ^i_jT}$ while preserving the rotation angle; in axis-angle format $o_t=(\hat{u}_t, \phi_t)$, the retargeted rotation delta is therefore $\tilde{o}_t=(R_{\Delta ^i_jT} \cdot \hat{u}_t, \phi_t)$.
The gripper command remains unchanged ($\tilde{g}_t = g_t$), yielding the retargeted action $\tilde{a}_t = (\tilde{p}_t, \tilde{o}_t, g_t)$.
Pairing the virtual RGB images with the retargeted proprioception and actions $\tilde{a}^j_t$ across the trajectory provides a synthetic demonstration at $^\mathit{b}_\mathit{w}\tilde{T}_j$.

\textbf{Constructing the Augmented Dataset.} For each of the $N_{\text{src}} = N \times M$ source manipulation demonstrations in $\mathcal{D}_{\text{src}}$, we generate $K$ synthetic demonstrations from $K$ sampled novel base poses, producing a synthetic dataset $\mathcal{D}_{\text{gen}}$ of $K \cdot N_{\text{src}}$ demonstrations that we combine with $\mathcal{D}_{\text{src}}$ to form $\mathcal{D} = \mathcal{D}_{\text{src}} \cup \mathcal{D}_{\text{gen}}$ for policy training (Sec.~\ref{sec:imitation_learning}).

\subsection{Imitation Learning}
\label{sec:imitation_learning}

We train a visuomotor policy on the union of source demonstrations $\mathcal{D}_{\text{src}}$ and synthetic demonstrations $\mathcal{D}_{\text{gen}}$ using Diffusion Policy~\cite{chi2023diffusionpolicy}, which models the conditional action distribution as a denoising diffusion process and is trained by minimizing the standard noise prediction loss:
\begin{equation}
    \mathcal{L} = \mathbb{E}_{k, \epsilon, (o, a) \sim \mathcal{D}} \left[ \| \epsilon - \epsilon_\theta(o, a^k, k) \|^2 \right],
\end{equation}
where $a^k$ is the ground-truth action sequence noised at denoising step $k$ with Gaussian noise $\epsilon$. The policy conditions on the RGB observation and end-effector proprioception, and outputs an action sequence of end-effector pose deltas. To balance the two datasets during training, we co-train with a sampling ratio of $0.25$ for $\mathcal{D}_{\text{src}}$ and $0.75$ for $\mathcal{D}_{\text{gen}}$, anchoring the policy to the source data while leveraging the broader pose coverage of the synthetic data.

%% file: sections/6_experiment.tex
\section{Experimental Evaluation}

We evaluate \methodname{} on five \emph{real-world} mobile manipulation tasks across three scenes, comparing against four baselines: no augmentation and three augmentation methods. Our experiments focus on a) whether \methodname{} improves policy success at unseen base poses, b) how performance varies with the distance between the deployment-time base pose and the closest demonstration pose, and c) how policy success changes with the number of source demonstrations available for augmentation. Additionally, we evaluate the quality
of the generated data itself. This requires ground truth observations at poses where no
demonstration was collected, which is impractical in the real world. A human
teleoperator produces a different motion on every attempt, so even a re-collected
trajectory cannot serve as ground-truth for the generated demonstration. 
Therefore, we develop one task in \emph{simulation}, where camera and object poses are
known and source trajectories can be replayed from arbitrary base poses.

\begin{figure}[!t]
\centering
\vspace*{0.5\baselineskip}
\begin{subfigure}{\linewidth}
    \centering
    \settoheight{\sceneimgheight}{%
        \includegraphics[width=0.235\linewidth]{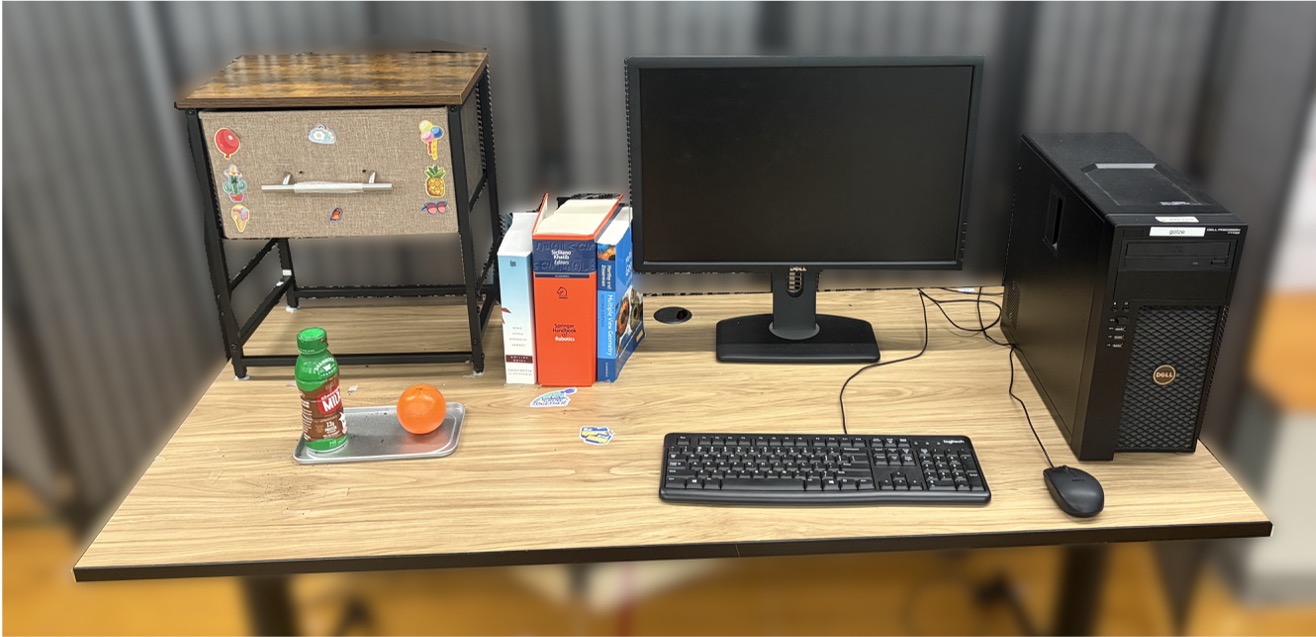}}%
    \includegraphics[width=0.235\linewidth]{figures/task_fig_images/desk_cropped.jpg}\hfill
    \includegraphics[width=0.235\linewidth]{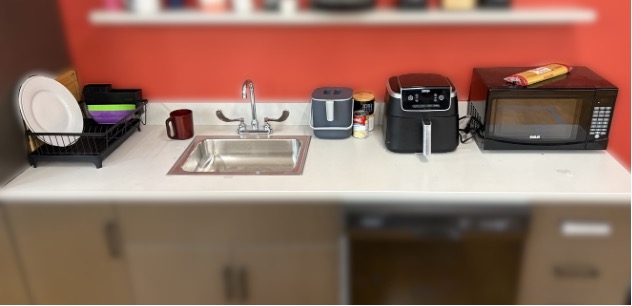}\hfill
    \includegraphics[width=0.235\linewidth]{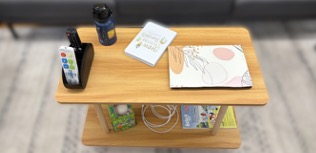}\hfill
    \begin{tikzpicture}[inner sep=0pt, outer sep=0pt]
        \draw[dashed, line width=1pt, black] (0,0) -- (0,\sceneimgheight);
    \end{tikzpicture}\hfill
    \includegraphics[width=0.235\linewidth]{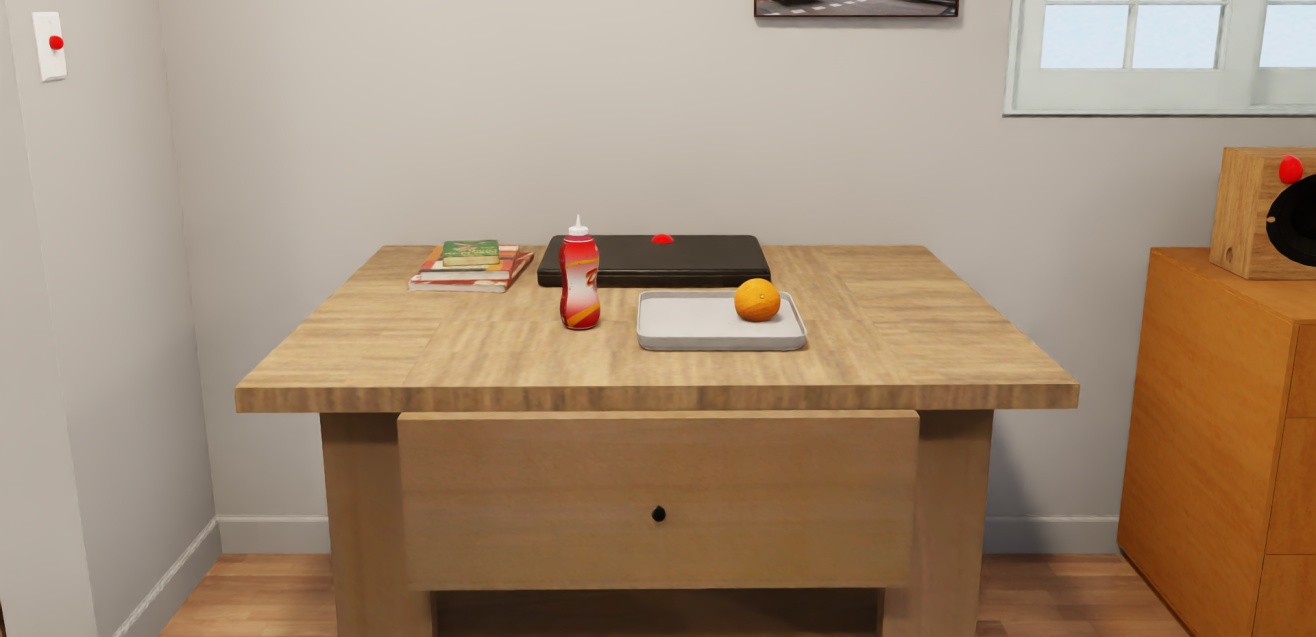}
    \caption{Scenes: desk, kitchen, and living room environments used in our experiments;
    the simulated scene is shown at right.}
\end{subfigure}%
\vspace{0.25em}
\begin{subfigure}{\linewidth}
    \centering
    \settoheight{\taskimgheight}{%
        \includegraphics[width=0.157\linewidth]{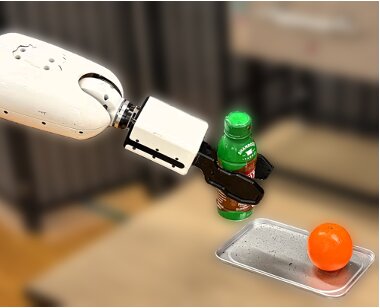}}%
    \includegraphics[width=0.157\linewidth]{figures/task_fig_images/milk_bottle_cropped.jpg}\hfill
    \includegraphics[width=0.157\linewidth]{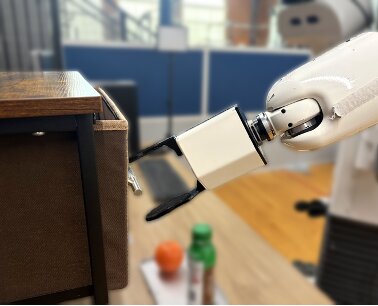}\hfill
    \includegraphics[width=0.157\linewidth]{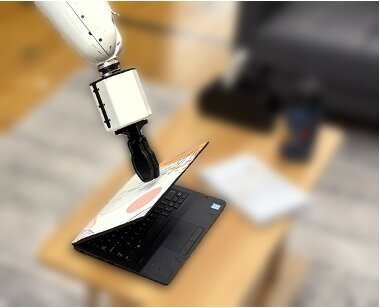}\hfill
    \includegraphics[width=0.157\linewidth]{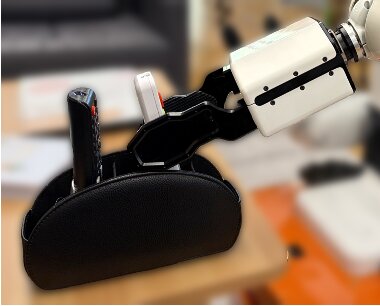}\hfill
    \includegraphics[width=0.157\linewidth]{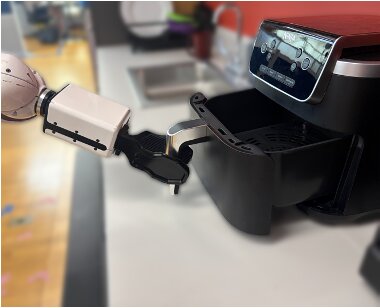}\hfill
    \begin{tikzpicture}[inner sep=0pt, outer sep=0pt]
        \draw[dashed, line width=1pt, black] (0,0) -- (0,\taskimgheight);
    \end{tikzpicture}\hfill
    \includegraphics[width=0.157\linewidth]{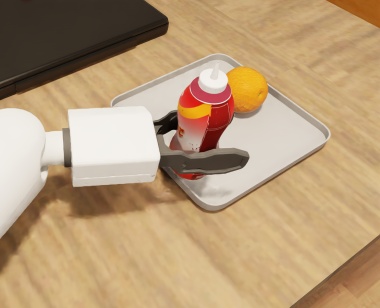}
    \caption{Tasks: placing a bottle, opening a drawer, closing a laptop, picking up a remote,
    and opening an air fryer in the real world; placing a bottle in simulation.}
\end{subfigure}
\caption{\textbf{Tasks and scenes used in our experiments.} We evaluate \methodname{} on five
real-world manipulation tasks across three scenes (top: scenes; bottom: per-task close-ups),
covering manipulation of both rigid and articulated objects. The simulated scene and its task,
\textsc{Sim Pick-Place Bottle} (right of the dashed lines), are used to measure the fidelity of
\methodname{}'s generated demonstrations against ground truth.}
\label{fig:tasks}
\end{figure}

\begin{figure*}[!ht]
    \centering
    \vspace*{0.5\baselineskip}

    
        \centering
         \includegraphics[width=1.0\textwidth]{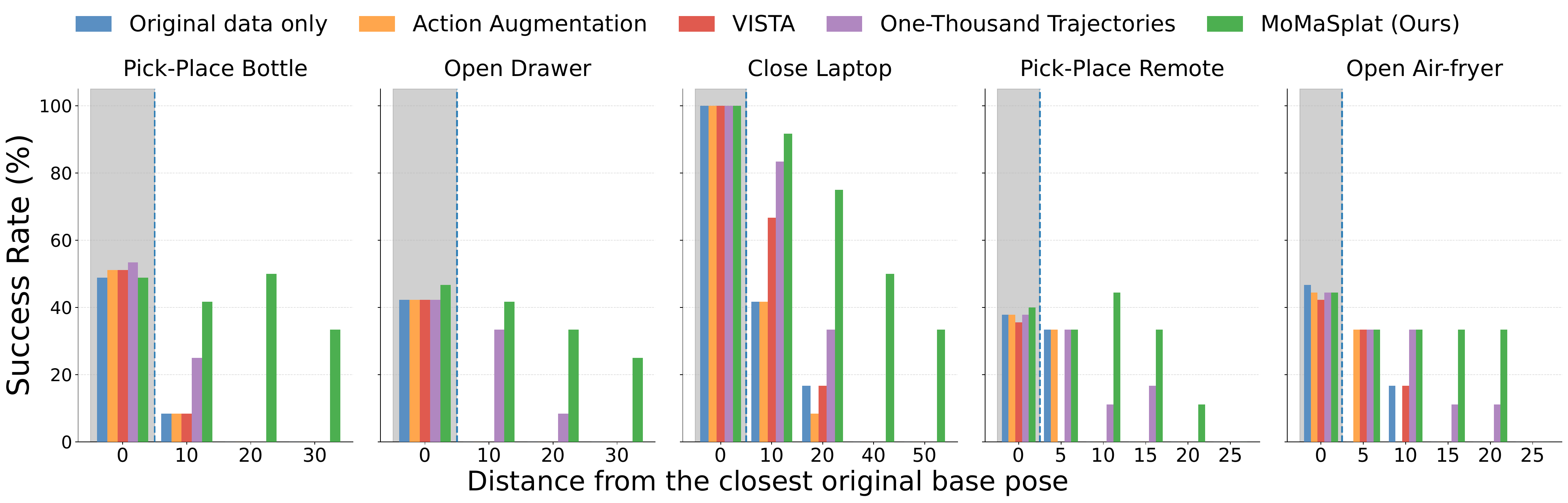}

\caption{\textbf{Policy success across base poses.}
For each task, we evaluate 12 unseen base locations distributed across categorical distance bins that vary per task, with 3 trials per pose, and report success rate by distance to the nearest demonstration pose. The shaded gray region denotes the original base poses; locations 5--50\,cm away are unseen deployment poses. \methodname{} maintains substantially higher success as the base pose shifts away from the source demonstrations (where \emph{Source Only} is labeled ``Original data only'').}
    \label{fig:succes_rates}

\end{figure*}

\begin{figure*}[!ht]
    \centering

    
        


    \begin{subfigure}{\textwidth}
        \centering
         \includegraphics[width=1.0\linewidth]{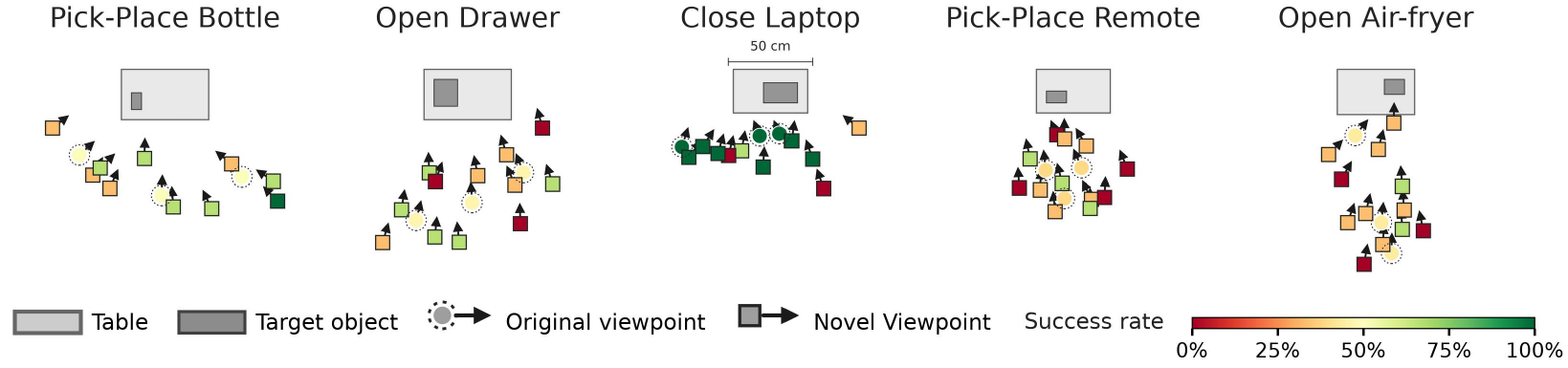}
       
        \label{fig:task_views}
    \end{subfigure}
\caption{\textbf{Spatial distribution of policy evaluation poses.}
Top-down view of each task showing the table and target-object locations, the original demonstration base poses, and the test base poses used for policy evaluation. Original demonstration poses are marked with circles, while evaluation poses are marked with squares; arrows indicate the robot heading at each pose. Evaluation poses are colored by \methodname{} policy success rate, providing spatial context for the distance-based results in Fig.~\ref{fig:succes_rates}.}
    \label{fig:eval_layout}
    \vspace{-1.2em}
\end{figure*}
\subsection{Experimental Setup}
\label{sec:exp_setup}

\textbf{Tasks.} The robot performs (i) \textsc{Pick-Place Bottle} and (ii) \textsc{Open Drawer} in the \emph{desk} scene; (iii) \textsc{Pick-Place Remote} and (iv) \textsc{Close Laptop} in the \emph{living room}; and (v) \textsc{Open Air-fryer} in the \emph{kitchen} (Fig.~\ref{fig:tasks}). We additionally build (vi) \textsc{Sim Pick-Place Bottle} in OmniGibson~\cite{li2023behavior}, a simulated scene modeled after the \emph{desk} environment that follows the same task structure as (i) without replicating its exact geometry. The tasks span rigid pick-place and single-DoF articulated manipulation. We collect $75$ manipulation demonstrations for \textsc{Close Laptop} and $90$ for each other task across three base poses, forming $\mathcal{D}_{\text{src}}$. We additionally collect nine navigation trajectories $\mathcal{D}_{\text{nav}}$ from different starting poses to the target base poses.

\textbf{Evaluation Protocol.} For each task, we deploy the trained policy at $12$ unseen base poses across the scene, with heading centered on the object, and perform $3$ rollouts per pose.

\textbf{Baselines.} We compare against four baselines.
\emph{Source Only} (labeled ``Original data only'' in Fig.~\ref{fig:succes_rates}) trains Diffusion Policy~\cite{chi2023diffusionpolicy}
without augmentation. \emph{Action Augmentation}~\cite{florence2019self}
adds noise to proprioceptive states and actions while preserving
observations. \emph{VISTA}~\cite{tian2024view} generates novel-view
images without transforming actions.
\emph{One-Thousand Trajectories}~\cite{pan2025one} reconstructs the
static scene using 3DGS; we additionally provide our action
transformation to account for changes in the robot base frame.
All methods train on the same source demonstrations. For
\textsc{Pick-Place Bottle}, we restrict synthetic generation for
\emph{One-Thousand Trajectories} to demonstrations whose initial
object pose matches the reconstructed scene, since its static
reconstruction cannot represent variation in this pose.


\textbf{Data Augmentation Setup.} Each method generates 10 synthetic demonstrations per source demonstration, yielding 750--900 synthetic demonstrations per task except \emph{One-Thousand Trajectories} on \textsc{Pick-Place Bottle}, which generates 400. For \methodname{} and \emph{One-Thousand Trajectories}, we sample base poses within a $1.2\times1.2$~m region around each source pose (Sec.~\ref{sec:camera_sampling}). Frame alignment uses 50 frames from one navigation demonstration.

\subsection{Results}
\label{sec:results}
\textbf{Performance at Unseen Base Poses.}
Our main experimental results for policy success using \methodname{} are summarized in
Fig.~\ref{fig:succes_rates}, where success rates are binned by their minimum distance to the nearest demonstration pose. We additionally evaluate the 3 original demonstration base poses with 15 rollouts per pose. At the original base poses, all methods perform comparably. The picture changes sharply at unseen poses. When we average the results over all five tasks, \methodname{} succeeds on 37\% of unseen evaluations, more than twice the rate of the strongest baseline (\emph{One-Thousand Trajectories} at 17\%) and four to six times that of the remaining baselines (\emph{Source Only} 6\%, \emph{VISTA} 8\%, \emph{Action Augmentation} 6\%). Critically, \emph{Source Only} and \emph{Action Augmentation} collapse to near-zero success beyond 10~cm on most tasks, confirming that perturbing actions without changing the observation gives little robustness to base-pose shifts. \emph{VISTA} performs poorly throughout; since it is designed for static-base settings, it does not transform actions to match the synthesized viewpoints. Moreover, its lack of temporal consistency produces jittery camera trajectories that further degrade the augmented data. \emph{One-Thousand Trajectories} degrades more gracefully but still falls off at large distances, since it augments only the free-space approach motion. \methodname{} achieves the highest overall unseen-pose success rate on every task, outperforming or matching all baselines across every unseen distance bin while sustaining substantial performance at the largest evaluated distances. For example, \methodname{} achieves 33\% on \textsc{Pick-Place Bottle} at 30~cm and 50\% on \textsc{Close Laptop} at 40~cm, where every baseline has dropped to zero success rates.
Finally, to better contextualize the geometry of our tasks and scenes, Fig.~\ref{fig:eval_layout} shows the spatial layout of these evaluation locations across each workspace.

 \textbf{Effect of Number of Source Demonstrations.} To evaluate how the number of source demonstrations affects \methodname{}, we use the \textsc{Pick-Place Bottle} task and train policies using 10, 50, and 90 source demonstrations. For each source demonstration, we generate 10 synthetic demonstrations. Across four test locations at a distance of $10$--$20$~cm from the demonstration poses, we observe that success degrades gracefully with fewer source demonstrations. Specifically, we achieve success rates of $33\%$, $42\%$, and $50\%$ for 10, 50, and 90 source demonstrations, respectively. For reference, training on the full set of 90 source demonstrations without augmentation yields only an $8\%$ success rate at the same test poses. These results show that \methodname{} substantially improves base-pose robustness even with a limited number of source demonstrations.

\textbf{Generation Quality.} We first evaluate generation quality in simulation, where the ground-truth counterpart of a generated demonstration is obtained by executing its transformed actions in the simulator and rendering the on-board camera at every timestep. 
We use the \textsc{Sim Pick-Place Bottle} task and evaluate $75$ generated demonstrations, taking every tenth frame for a total of $2{,}486$ frames. 
Unlike the policy evaluations above, which measure distance between base poses, here we measure how far each generated view is from the views that were used to build the reconstruction.
For the $75$ demonstrations, we find this distance to be between $0.4$ and $27.3$~cm.
%
Since the static reconstruction does not change over a demonstration, we score the scene separately by rendering the scene splat and the empty simulated scene from the same camera pose once per demonstration, rather than 
considering
every frame. We report PSNR, SSIM, and LPIPS over the full image, the scene, and over the robot and object regions using the simulator's segmentation masks.  Table~\ref{tab:generation_quality} reports the detailed results.
The reconstructed scene achieves the highest fidelity at $26.40$~dB, followed by the manipulated object at $18.18$~dB and the robot at $17.43$~dB. For reference, the original 3DGS work~\cite{kerbl20233d} reports roughly $23$~dB PSNR on Tanks \& Temples~\cite{10.1145/3072959.3073599}. This places our scene reconstruction in the range of standard novel-view synthesis, while the lower object PSNR reflects the harder setting of reconstructing a moving object with limited viewpoint coverage. The robot error is mainly caused by lighting and shadow differences introduced during compositing. We also evaluate object-pose accuracy to measure how well our estimated object trajectory matches the true object motion 
of the observations. On $30$ source demonstrations, including poses directly registered by VGGT and interpolated poses, the predicted object pose has a mean translation error of $1.46 \pm 0.71$~cm and rotation error of $4.01^\circ \pm 4.66^\circ$ relative to the simulator ground truth. 


On real data, we evaluate scene reconstruction by masking out the robot and the manipulated object, and we obtain $24.00$~dB PSNR on $110$ held-out frames. For the manipulated object, we use its bounding-box crop and we mask out the background. At poses directly registered by VGGT, we obtain $23.63$~dB, while at interpolated poses we obtain $22.59$~dB. We do not report full-image view-synthesis metrics on real data because the simulated robot differs in appearance from the real robot and this appearance mismatch dominates the pixel-wise error that the view synthesis metrics evaluate.

Overall, the simulation results show that \methodname{} produces observations that closely match the ground truth, while the real-world results show high-fidelity scene reconstruction and consistent object reconstruction across both directly estimated and interpolated poses.

\begin{table}[t]
\vspace{\intextsep}
\centering
\scriptsize
\setlength{\tabcolsep}{6pt}
\begin{tabular}{lccc}
\toprule
Region & PSNR (dB) $\uparrow$ & SSIM $\uparrow$ & LPIPS $\downarrow$ \\
\midrule
Full image & $21.27 \pm 1.12$ & $0.902 \pm 0.019$ & $0.253 \pm 0.030$ \\
Scene      & $26.40 \pm 2.18$ & $0.936 \pm 0.021$ & $0.248 \pm 0.037$ \\
Object     & $18.18 \pm 2.15$ & $0.688 \pm 0.081$ & $0.341 \pm 0.106$ \\
Robot      & $17.43 \pm 1.30$ & $0.812 \pm 0.057$ & $0.287 \pm 0.048$ \\
\bottomrule
\end{tabular}

\caption{\textbf{Generated-observation fidelity against simulator ground truth.} Mean $\pm$ std over $75$ generated demonstrations spanning viewpoints with distances of $0.4$--$27.3$~cm from the reconstruction views. \emph{Full image}, \emph{object}, and \emph{robot} use every tenth frame ($2486$ total frames), with object and robot restricted to their masks; \emph{scene PSNR} evaluation uses one frame per demonstration ($75$ in total).}



\label{tab:generation_quality}
\end{table}

%% file: sections/7_conclusions.tex
\section{Conclusion and Limitations}
\textbf{Conclusion.} We propose \methodname{}, a framework for novel-view augmentation of mobile manipulation demonstrations that reconstructs the static scene and manipulated object in 3D, then re-renders full demonstrations, including the post-contact interaction phase, from unseen base poses. Unlike prior NVS augmentation methods that model only the static scene, \methodname{} produces geometrically consistent observations across the entire trajectory. Across five real-world tasks in three scenes, this improves policy robustness, outperforming prior augmentation methods and maintaining performance beyond the distances where baselines fail. We will publicly release our code to support further research.

\textbf{Limitations.} \methodname{} assumes mostly static scenes and rigid, visible objects, making deformable objects, heavy occlusions, and texture-poor surfaces challenging. Furthermore, because $\mathcal{G}_{\text{scene}}$ is reconstructed from pre-interaction navigation views with object masks excluded, regions occluded by the object in its initial state (such as the surface beneath a lifted bottle or the interior of an opened drawer or air-fryer basket) are never observed during scene reconstruction and remain unmodeled when disoccluded. It also depends on consistent open-world segmentation and produces non-physical demonstrations by replaying source object trajectories without modeling contact forces or dynamics. Finally, view quality depends on navigation-data coverage. Addressing dynamic backgrounds, deformable objects, physical contact, and cluttered scenes remains future work.

%% file: sections/9_appendix.tex
\label{s:appendix}

\subsection{Baseline Details}

In this section, we provide more details about the baselines we are using in our evaluation.

\textbf{Source Only.} This baseline trains the same Diffusion Policy~\cite{chi2023diffusionpolicy} as we do, but using only the source demonstrations $\mathcal{D}_{\text{src}}$, without any augmentation. This allows us to quantify how well a policy trained only on the original base poses generalizes to unseen poses.

\textbf{Action Augmentation.} Following~\cite{florence2019self}, we augment each demonstration by perturbing the proprioceptive state and the action labels with zero-mean Gaussian noise while leaving the image observations unchanged. This exposes the policy to small action variations, but because the images remain the same, this augmentation provides no information about how the scene looks from a different base pose. We use this data to train the same Diffusion Policy and use the same training schedule as we do with all other baselines.

\textbf{VISTA.} VISTA~\cite{tian2024view} augments a demonstration by re-rendering its observations from perturbed camera viewpoints with ZeroNVS~\cite{sargent2024zeronvs}, a zero-shot single-image novel-view-synthesis diffusion model, while leaving the actions unchanged. We run the publicly released ZeroNVS model and checkpoint.\footnote{\url{https://github.com/s-tian/ZeroNVS}} For each source demonstration, we sample a small set of viewpoint perturbations: a planar camera offset $(dx, dy)$ drawn uniformly within $\pm 10$~cm on each axis (the first sample is the unperturbed view), together with a single per-demonstration yaw that re-aims the camera at the object centroid, clipped to $\pm 0.5$~rad; beyond this perturbation range, we observe large artifacts in image generation and worse imitation learning results. We then process each frame independently: conditioning on the original egocentric RGB image (center-cropped and resized to $256 \times 256$) and the relative perturbation, ZeroNVS synthesizes the frame from the novel viewpoint. We sample with DDIM ($50$ steps, $\eta = 0$), a classifier-free guidance scale of $7.5$, and a $70^\circ$ field of view matching our camera. Because ZeroNVS is a single-image model applied frame by frame, the synthesized views are not temporally consistent across a trajectory. The generated frames are paired with the original, unmodified actions and are used to train the same Diffusion Policy as all other methods.

\textbf{One-Thousand Trajectories.}
One-Thousand Trajectories~\cite{pan2025one} reconstructs the static scene as a
3D Gaussian Splatting model and re-renders observations from novel
viewpoints. In our implementation, we reconstruct the static scene using the same pipeline and navigation data $\mathcal{D}_{\text{nav}}$ as $\mathcal{G}_{\text{scene}}$ in MoMaSplat, but without masking out the manipulated object, so that the static scene model contains the manipulated object at its initial pose.

We use its novel-view augmentation component without the additional
trajectory augmentation proposed in~\cite{pan2025one}. To account for changes in
the robot base frame, we apply the same delta-based action retargeting
used by MoMaSplat (Sec.~\ref{sec:demo_synthesis}). This baseline relies on a static scene
reconstruction rather than explicitly reconstructing the dynamic
manipulated object. For \textsc{Pick-Place Bottle}, we restrict
synthetic generation to the 40 demonstrations whose initial object
pose matches the reconstructed scene, producing 400 synthetic
demonstrations. All 90 source demonstrations remain in the
policy-training set. For the other tasks, the initial object pose
is fixed, so all source demonstrations are used for synthetic
generation.

\subsection{Method Details}

\subsubsection{3D Scene Model Reconstruction}

\textbf{Structure from motion and Gaussian Splatting.} We reconstruct the static scene from the navigation frames $\mathcal{D}_{\text{nav}}$. We run structure-from-motion using COLMAP~\cite{schonberger2016structure} with a sequential matcher over the navigation frames, together with a small set of context frames sampled just before each manipulation. This step returns per-image camera poses and a sparse point cloud. The sparse points are used to initialize a 3D Gaussian Splatting reconstruction that we train with WildGaussians~\cite{kulhanek2024wildgaussians}\footnote{\url{https://github.com/jkulhanek/wild-gaussians}}. We use the \texttt{nerfbaselines} framework and train for $30{,}000$ iterations. WildGaussians augments 3DGS with per-image appearance and exposure embeddings and a DINOv2-based uncertainty head that down-weights transient or inconsistent regions, such as the robot or moving objects, during optimization. We use a DSSIM weight of $0.2$ in the photometric loss and hold out every eighth frame for evaluation.

\textbf{Metric scaling.} Structure-from-motion recovers the scene only up to an unknown global scale. However, our action retargeting is metric, so we rescale the reconstruction using the robot's onboard depth sensor measurements. For each navigation frame we render a depth map from the scene model and compare it to the measured depth from the head-mounted RGB-D camera (stored in millimeters and converted to meters). We consider only pixels with valid measured depth (in $(0, 9]$~m, at least $1000$ per frame) and for each frame, we take the median of the per-pixel ratios of measured to rendered depth. Eventually, the scale factor is set as the median of these per-frame median values, making it robust to outliers. This factor allows us to convert rendered depth and Gaussian coordinates into meters.

\textbf{Scene pose interpolation.} For the navigation frames that have not been registered by COLMAP, we estimate their scene camera poses using the same strategy as we do with objects (see Sec.~\ref{sec:object_model_supmat}). This includes SLERP-based interpolation, smoothness filtering, and cyclic-outlier rejection, applied per demonstration.

\subsubsection{3D Object Model Reconstruction}
\label{sec:object_model_supmat}
\textbf{Segmentation.} To obtain the per-frame object masks $\mathcal{M}_t^j$, we use Grounded-SAM~\cite{ren2024grounded}, prompting an open-world detector, GroundingDINO~\cite{liu2023grounding}, with a per-object text label to localize the manipulated object, and propagating the detected box through the clip using SAM~2~\cite{ravi2024sam2}\footnote{Grounded-SAM~2: \url{https://github.com/IDEA-Research/Grounded-SAM-2}}. We use a forward and a reverse pass from a per-object seed frame. When multiple objects are tracked, pixels claimed by more than one object are removed from all of them (exclusive masks). Object reconstruction uses the per-object masks to exclude the
background and robot. For scene reconstruction, we additionally
obtain a robot mask by rendering the robot in the simulator at
the recorded configuration and camera pose. We exclude the union
of the object masks and this robot mask when training
$\mathcal{G}_{\text{scene}}$.

\textbf{Frame Selection Strategy.} A source manipulation demonstration $d^j_{\text{src}} \in \mathcal{D}_{\text{src}}$ contains several hundred RGB frames recorded at video rate, far more than the feed-forward model can process. Simultaneously, near-duplicate frames add little to the pose estimate while inflating compute. We therefore subsample each demonstration to a compact, motion-diverse set of roughly $10$--$20$ frames before passing it to the model, using the per-frame object masks $\mathcal{M}_t^j$.

We first discard frames whose object mask is small or poorly segmented, removing views where the object is distant or occluded. From the remaining frames we drop near-duplicate, stationary views: scanning in temporal order, we keep a frame only when its mask bounding-box IoU with the previously kept frame falls below $0.8$ (the same threshold for all tasks), or when more than $20$ frames have elapsed since the last kept frame. The first and last frame of every demonstration are always retained to anchor the reconstruction to the endpoints of the clip. Additionally, $50$ frames from one navigation demonstration are included for frame alignment, and the combined set is capped at the feed-forward model's $400$-frame input budget.

\textbf{Bundle Adjustment.} 
For our feed-forward model,
we use VGGT~\cite{wang2025vggt}\footnote{\url{https://github.com/facebookresearch/vggt}}, as implemented by the training-free FastVGGT~\cite{shen2025fastvggt} acceleration\footnote{\url{https://github.com/mystorm16/FastVGGT}}. For each selected frame, it predicts an object-relative camera pose $\hat{T}_{o,t}$ together with a per-frame intrinsics matrix and a dense depth map. These predictions are accurate but not multi-view consistent, so we refine them with a bundle-adjustment step. We first unproject the predicted depth into a set of 3D points and predict dense 2D feature tracks across the selected frames. Initializing a reconstruction with the predicted poses, intrinsics, and points, we filter tracks by reprojection error, using an adaptive threshold that grows until a sufficient fraction of frames retain enough inliers, and by track visibility, requiring every 3D point to be observed in at least two frames. We then run bundle adjustment, jointly optimizing the camera poses and 3D points (and, when intrinsics are not supplied, the focal length and principal point) to minimize reprojection error. The refined, multi-view-consistent poses replace $\hat{T}_{o,t}$ as the input to the $\mathrm{Sim}(3)$ alignment.

\textbf{Frame Alignment.} The poses estimated above lie in the object model's internal frame $v$ with arbitrary scale, orientation, and origin, whereas the scene lives in the world frame $w$. We align them using the navigation frames, where the object is static so its recovered trajectory is explained purely by the camera motion. Effectively, the same cameras are posed twice: by the object model in $v$ and by structure-from-motion in $w$. Matching the navigation cameras by frame, we fit the similarity transform $S: v \rightarrow w$ from their camera centers with the Umeyama algorithm~\cite{umeyama1991least}, solving for scale, rotation, and translation in closed form. To be robust to mis-registered frames, we wrap Umeyama in RANSAC: we repeatedly fit $S$ from minimal $5$-camera subsets, reject hypotheses whose scale leaves a plausible range, score each by the number of camera centers whose residual is below $0.2$~m, keep the highest-consensus hypothesis, and refit on its inliers followed by a Huber-robust nonlinear refinement restricted to those inliers. We report the alignment quality as the camera-center residual (mean and median) and the RMS of the aligned sparse-point residuals. Applying $S$ maps the object poses into the world frame, $T_{o,t} = S \cdot \hat{T}_{o,t}$.

\textbf{Object Gaussian Optimization.} Given the registered frames and their aligned poses $T_{o,t}$, we fit a 3D Gaussian Splatting model~\cite{kerbl20233d} $\mathcal{G}_{\text{obj}}$ of the object with the gsplat~\cite{ye2025gsplat} trainer\footnote{\url{https://github.com/nerfstudio-project/gsplat}} for $60{,}000$ steps, supervising only the segmented object through its RGBA images, where the alpha channel is the object mask composited over a white background. We minimize an $L_1$ plus SSIM photometric loss (SSIM weight $0.2$) together with an $L_1$ term on the rendered opacity against the mask. The per-frame poses are optimized jointly with the Gaussian parameters at a small learning rate of $10^{-5}$, so the photometric loss is backpropagated through both appearance and pose, yielding the pose-disentangled model $\mathcal{G}_{\text{obj}}$ that can be rendered at arbitrary 6-DoF poses.

\textbf{Pose Interpolation.} Because of frame selection, the feed-forward model only returns camera poses at the selected keyframes. In contrast to that, synthesizing a demonstration requires an object pose $T_{o,t}$ at \emph{every} timestep of the source trajectory. We therefore densify the keyframe poses by interpolation: rotations are interpolated with spherical linear interpolation (SLERP) and translations are interpolated linearly, between consecutive keyframes. To suppress residual estimation noise we add three refinements before interpolating. \emph{(i)} A smoothness filter drops keyframes whose translation and rotation relative to the previous keyframe both fall below small thresholds, removing jitter from near-stationary stretches. \emph{(ii)} A cyclic-outlier filter removes backtracking $A\!\rightarrow\!B\!\rightarrow\!A$ keyframes, where an excursion to $B$ is large enough to be meaningful yet the trajectory returns close to $A$, a signature of transient pose-estimation errors. \emph{(iii)} An ease-in parameterization places interpolated samples so that pose changes are concentrated toward the end of each keyframe interval, better matching the observed motion profile than uniform piecewise-linear filling. The densified trajectory $\{T_{o,t}\}$ is what we rasterize $\mathcal{G}_{\text{obj}}$ along when compositing novel-view observations.

\subsection{Base Pose Sampling Strategy} Here, we expand on the procedure used to draw the sampled base poses $\mathcal{T}_i^{\text{sampled}} = \{\tilde{T}_1, \dots, \tilde{T}_{K}\}$ around each original base pose $T_i$ (Sec.~\ref{sec:demo_synthesis}). Sampling operates per source demonstration about an anchor pose recovered for that demonstration, i.e., the original base position and heading expressed in the world frame, as it is established by the scene reconstruction.

\emph{Planar position.} We draw the planar offset $(\tilde{x}_j, \tilde{y}_j)$ from a distribution centered at the anchor. By default this is uniform over a square region around the original base pose (the region size used in our experiments is reported in Sec.~\ref{sec:exp_setup}); we also support an isotropic Gaussian with magnitude bounds for annulus-shaped coverage.

\emph{Heading.} For each sampled position we set the base heading $\tilde{\theta}_j$ so that the egocentric camera points at the object's 3D centroid (accounting for the camera's fixed offset from the base), then perturb it with uniform noise (default $\pm 20^\circ$) so the policy sees the object at a range of positions in the camera view. The robot head is fixed, so the object is kept in view through the base pose alone; we discard samples that would place the object near the image edge or out of view before running the costlier IK check.

\emph{IK feasibility.} Following Sec.~\ref{sec:demo_synthesis}, a sampled base pose is retained only if the source arm trajectory, with the end effector held at the same pose relative to the manipulated object, admits a valid inverse-kinematics solution. We verify reachability at several probe timesteps spanning the demonstration and accept the sample only if all of them are reachable, querying the curobo inverse-kinematics solver for the TIAGo robot inside the OmniGibson simulator~\cite{li2023behavior}\footnote{OmniGibson: \url{https://github.com/StanfordVL/OmniGibson}; cuRobo: \url{https://github.com/NVlabs/curobo}}. Infeasible candidates are rejected and resampled, up to a fixed draw budget per demonstration, and the first $K$ accepted samples are kept.

\begin{table}[h]
\centering
\scriptsize
\setlength{\tabcolsep}{4pt}
\begin{tabular}{lll}
\toprule
Stage & Parameter & Value \\
\midrule
Scene SfM      & matcher                    & sequential (COLMAP) \\
Scene 3DGS     & method / iterations        & WildGaussians / $30{,}000$ \\
Scene 3DGS     & DSSIM weight               & $0.2$ \\
Metric scale   & valid-depth range          & $(0, 9]$~m, $\geq 1000$~px/frame \\
Segmentation   & detector / tracker         & GroundingDINO / SAM2 \\
Frame selection& IoU thresh.\ / max gap     & $0.8$ / $20$ frames \\
Frame selection& VGGT input budget          & $400$ frames per object \\
Object BA      & visibility per 3D point    & $\geq 2$ views \\
$\mathrm{Sim}(3)$ RANSAC & subset / trials / inlier & $5$ / $8000$ / $0.2$~m \\
Object 3DGS    & steps / SSIM weight        & $60{,}000$ / $0.2$ \\
Object 3DGS    & pose-opt learning rate     & $10^{-5}$ \\
Base sampling  & region / yaw noise         & $120 \times 120$~cm / $\pm 20^\circ$ \\
Generation     & synthetic per source demonstration & $10$ \\
Policy         & obs / action horizon       & $2$ / $8$ \\
Policy         & DDIM train timesteps       & $50$ \\
Policy         & optimizer / learning rate  & AdamW / $3 \times 10^{-4}$ \\
Policy         & batch size / training steps & $32$ / $100{,}000$ \\
Policy         & source : generated ratio   & $0.25 : 0.75$ \\
\bottomrule
\end{tabular}
\caption{Key hyperparameters across the M\textsc{o}M\textsc{a}S\textsc{plat} pipeline.}
\label{tab:hyperparams}
\end{table}

\subsection{Imitation Learning} We provide implementation details for the policy trained with Diffusion Policy~\cite{chi2023diffusionpolicy} on the union of source and generated demonstrations (Sec.~\ref{sec:imitation_learning}).

\emph{Architecture.} The policy is a conditional 1D U-Net denoiser. The egocentric head-camera RGB image is encoded by a ResNet-18 trained from scratch followed by a spatial-softmax pooling layer ($32$ keypoints), and the resulting visual features are concatenated with the low-dimensional proprioception to form the conditioning vector.

\emph{Observations and actions.} At each step the policy conditions on the egocentric head-camera RGB image (resized to $224 \times 224$) and the end-effector proprioception, stacked over the most recent $n_{\text{obs}} = 2$ steps. It predicts a chunk of $n_{\text{act}} = 8$ future actions, each a $7$-dimensional command consisting of an end-effector translation delta, a rotation delta, and a gripper command expressed in the robot base frame, normalized to zero mean and unit variance. During training we apply random crop and translation together with color-jitter augmentation to the RGB input.

\emph{Diffusion process.} We use a DDIM noise scheduler with $50$ training timesteps, a cosine-based $\beta$ schedule, and $\epsilon$ (noise) prediction, optimizing the mean-squared denoising loss of Eq.~(1). At inference, actions are produced by running the reverse process for the same number of denoising steps.

\emph{Optimization.} We optimize with AdamW (learning rate $3\times10^{-4}$, weight decay $10^{-4}$) under a cosine schedule with $1000$ warm-up steps and a batch size of $32$ for $100{,}000$ gradient steps.

\emph{Source and synthetic co-training.} We train on the union of the source demonstrations $\mathcal{D}_{\text{src}}$ and the generated demonstrations $\mathcal{D}_{\text{gen}}$ using a weighted sampler that draws each training sequence from $\mathcal{D}_{\text{src}}$ with probability $0.25$ and from $\mathcal{D}_{\text{gen}}$ with probability $0.75$. This anchors the policy to the source data while exploiting the broader base-pose coverage of the synthetic data.

\emph{Hardware and data.} We use a TIAGo mobile manipulator with a head-mounted RGB-D camera ($70^\circ$ field of view). We collect 75 source demonstrations for \textsc{Close Laptop} and 90 for each other task across three base poses, and generate $10$ synthetic demonstrations per source demonstration within a $120 \times 120$~cm region around each source base pose. We evaluate each policy at $12$ unseen base poses with $3$ rollouts per pose. Table~\ref{tab:hyperparams} summarizes the main hyperparameters across the pipeline.